\documentclass[runningheads]{llncs}
\usepackage[T1]{fontenc}
\usepackage{graphicx,verbatim}
\usepackage{graphicx}
\usepackage{amsmath,amssymb}
\usepackage{booktabs}
\usepackage{multirow}
\usepackage{xcolor}
\usepackage{algorithm}
\usepackage{algorithmic}
\usepackage{subfigure}

\begin{document}

\title{Specializing Foundation Models via Mixture of Low‑Rank Experts for Comprehensive Head CT Analysis}

\titlerunning{Specializing Foundation Models via Mixture of Low‑Rank Experts}


\author{Youngjin Yoo\inst{1} \and
Han Liu\inst{1} \and
Bogdan Georgescu\inst{1} \and
Yanbo Zhang\inst{1} \and
Sasa Grbic\inst{1} \and
Michael Baumgartner\inst{1} \and
Thomas J. Re\inst{1} \and
Jyotipriya Das\inst{1} \and
Poikavila Ullaskrishnan\inst{1} \and
Eva Eibenberger\inst{2} \and
Andrei Chekkoury\inst{2} \and
Uttam K. Bodanapally\inst{4} \and
Savvas Nicolaou\inst{5} \and
Pina C. Sanelli\inst{6} \and
Thomas J. Schroeppel\inst{7} \and
Yvonne W. Lui\inst{3} \and
Eli Gibson\inst{1}}
\authorrunning{Yoo et al.}

%
\institute{Digital Technology and Innovation, Siemens Healthineers, Princeton, NJ USA \and
Department of Computed Tomography, Siemens Healthineers, Forchheim, Germany \and
Department of Radiology, New York University, New York, NY USA \and
Department of Radiology, University of Maryland Medical Center, Baltimore, MD USA \and
Department of Radiology, Vancouver General Hospital, Vancouver, BC Canada \and
Department of Radiology, Northwell Health, New York, NY USA \and
Department of Surgery, UCHealth Memorial Hospital, Colorado Springs, CO USA
}

\maketitle

\begin{abstract}
Foundation models pre-trained on large-scale datasets demonstrate strong transfer learning capabilities; however, their adaptation to complex multi-label diagnostic tasks—such as comprehensive head CT finding detection—remains understudied. Standard parameter-efficient fine-tuning methods such as LoRA apply uniform adaptations across pathology types, which may limit performance for diverse medical findings. We propose a Mixture of Low-Rank Experts (MoLRE) framework that extends LoRA with multiple specialized low-rank adapters and unsupervised soft routing. This approach enables conditional feature adaptation with less than 0.5\% additional parameters and without explicit pathology supervision. We present a comprehensive benchmark of MoLRE across six state-of-the-art medical imaging foundation models spanning 2D and 3D architectures, general-domain, medical-domain, and head CT–specific pretraining, and model sizes ranging from 7M to 431M parameters. Using over 70,000 non-contrast head CT scans with 75 annotated findings—including hemorrhage, infarction, trauma, mass lesions, structural abnormalities, and chronic changes-our experiments demonstrate consistent performance improvements across all models. Gains vary substantially: general-purpose and medical-domain models show the largest improvements (DINOv3-Base: +4.6\%; MedGemma: +4.3\%), whereas 3D CT-specialized or very large models show more modest gains (+0.2–1.3\%). The combination of MoLRE and MedGemma achieves the highest average detection AUC of 0.917. These findings highlight the importance of systematic benchmarking on target clinical tasks, as pretraining domain, architecture, and model scale interact in non-obvious~ways.


\end{abstract}

\section{Introduction}
Foundation models pre-trained on massive datasets have fundamentally shifted the paradigm of medical image analysis, demonstrating strong zero-shot and few-shot transfer capabilities across a wide range of tasks~\cite{ma2024medsam,kiraly2025medgemma,codella2024medimageinsight,hamamci2024foundation,zhu20253d,agrawal2025pillar,khan2025comprehensive,jiao2025foundation}. However, while these models achieve impressive performance on standardized benchmarks, their effectiveness in complex multi-label clinical scenarios—such as comprehensive non-contrast head CT analysis involving dozens of heterogeneous findings—remains insufficiently understood. Existing work largely focuses on single-organ detection or narrow subsets of abnormalities~\cite{titano2018automated,flanders2020construction}, leaving open questions about how foundation models handle the high-dimensional diagnostic complexity underlying automated radiology report generation~\cite{pons2016natural,hassanpour2016information,yang2023radiology}.

Adapting large foundation models to such specialized clinical tasks typically relies on parameter-efficient fine-tuning (PEFT)~\cite{ding2023parameter}, with Low-Rank Adaptation (LoRA)~\cite{hu2022lora,liu2025revisiting} emerging as a dominant approach. LoRA approximates weight updates through low-rank decomposition, enabling efficient adaptation with minimal additional parameters. However, standard LoRA applies a uniform low-rank update across all inputs, implicitly assuming that a single adaptation suffices for all pathology types. In comprehensive head CT analysis, this uniformity can lead to knowledge interference~\cite{ruder2017overview}, where features required to detect acute hemorrhage, chronic ischemia, trauma, and subtle structural abnormalities must compete for the same limited adaptation capacity.

Recent advances in mixture-of-experts architectures enable conditional, input-dependent specialization by activating different subnetworks for different inputs~\cite{jacobs1991adaptive,shazeer2017outrageously,fedus2022switch,lepikhin2020gshard}. Motivated by these principles, we propose \emph{Mixture of Low-Rank Experts (MoLRE)}, a parameter-efficient extension of LoRA that enables conditional adaptation via unsupervised soft routing with less than 0.5\% additional parameters. Our main contributions are as follows:
\begin{itemize}
    \item We introduce \emph{Mixture of Low-Rank Experts}, a conditionally routed, low-rank adaptation framework for foundation models.
    \item We present a large-scale benchmark of MoLRE across six 2D and 3D medical imaging foundation models (7M--431M parameters) on over 70{,}000 head CT scans with 75 neurological findings.
    \item We achieve state-of-the-art performance, with consistent gains across all models and a highest average AUC of \textbf{0.917} using MedGemma+MoLRE.
    \item We provide empirical insights showing that adaptation benefit depends on a complex interaction between pretraining domain, architecture, and model scale.
\end{itemize}

\section{Method}

\subsection{Mixture of Low-Rank Experts (MoLRE)}

MoLRE extends LoRA by introducing $K$ specialized low-rank experts with learned input-dependent routing. Given an input feature $\mathbf{x} \in \mathbb{R}^{d}$, the adapted output is computed as:

\begin{equation}
\mathbf{h} = \mathbf{W}_0\mathbf{x} + \sum_{i=1}^{K} g_i(\mathbf{x}) \cdot \Delta\mathbf{W}_i\mathbf{x}
\end{equation}

where $\mathbf{W}_0 \in \mathbb{R}^{d_{out} \times d}$ is the frozen pre-trained weight matrix, and $\Delta\mathbf{W}_i = \mathbf{B}_i\mathbf{A}_i$ represents the $i$-th expert's low-rank adaptation with $\mathbf{A}_i \in \mathbb{R}^{r \times d}$ and $\mathbf{B}_i \in \mathbb{R}^{d_{out} \times r}$, where $r \ll \min(d, d_{out})$ is the rank. The router network $g(\mathbf{x}): \mathbb{R}^{d} \rightarrow \mathbb{R}^{K}$ computes expert mixing weights via a two-layer MLP with softmax normalization:

\begin{equation}
\mathbf{g}(\mathbf{x}) = \text{Softmax}(\mathbf{W}_2 \cdot \text{ReLU}(\mathbf{W}_1\mathbf{x} + \mathbf{b}_1) + \mathbf{b}_2)
\end{equation}

where $\mathbf{W}_1 \in \mathbb{R}^{d_h \times d}$, $\mathbf{b}_1 \in \mathbb{R}^{d_h}$, $\mathbf{W}_2 \in \mathbb{R}^{K \times d_h}$, and $\mathbf{b}_2 \in \mathbb{R}^{K}$, with $d_h$ denoting the router hidden dimension. The softmax ensures $\sum_{i=1}^{K} g_i(\mathbf{x}) = 1$ and $g_i(\mathbf{x}) \geq 0$, enabling all experts to contribute with varying degrees based on the input.

\subsection{Integration with Foundation Models}
For 2D foundation models (e.g., DINOv3~\cite{simeoni2025dinov3}, MedGemma~\cite{kiraly2025medgemma}, MedImageInsight~\cite{codella2024medimageinsight}), we process volumetric medical scans with $S$ slices using slice-level feature extraction followed by MoLRE adaptation and attention-weighted pooling~\cite{ilse2018attention,liu2025revisiting,campanella2019clinical}. The router network $g(\mathbf{x})$ operates on individual slice features, enabling slice-specific expert selection, which is particularly beneficial for head CT where pathologies may be spatially localized. The complete inference pipeline for 2D foundation models is summarized in Algorithm~\ref{algo:2D_algo}.

\begin{algorithm}
\caption{MoLRE for 2D Vision Foundation Models}
\label{algo:2D_algo}
\begin{algorithmic}[1]
\REQUIRE Volumetric medical scan $\mathbf{X} \in \mathbb{R}^{B \times M \times S \times H \times W}$
\REQUIRE Foundation model $f_\theta$, MoLRE module $\text{MoLRE}_\phi$
\STATE Reshape: $\mathbf{X}' \leftarrow \text{reshape}(\mathbf{X}, [B \cdot S, M, H, W])$
\STATE Extract features: $\mathbf{F} \leftarrow f_\theta(\mathbf{X}')$ \COMMENT{$[B \cdot S, d]$}
\STATE Apply MoLRE: $\mathbf{F}' \leftarrow \text{MoLRE}_\phi(\mathbf{F})$ \COMMENT{$[B \cdot S, d]$}
\STATE Reshape: $\mathbf{F}' \leftarrow \text{reshape}(\mathbf{F}', [B, S, d])$
\STATE \textbf{Attention pooling:}
\STATE \quad $\mathbf{Q} \leftarrow \text{expand}(\mathbf{q}_{\text{learnable}}, B)$ \COMMENT{$[B, 1, d]$}
\STATE \quad $\alpha \leftarrow \text{Softmax}\left(\frac{\mathbf{Q} \mathbf{F}'^T}{\sqrt{d}}\right)$ \COMMENT{$[B, 1, S]$}
\STATE \quad $\mathbf{h} \leftarrow \alpha \mathbf{F}'$ \COMMENT{$[B, d]$}
\STATE Classify: $\mathbf{y} \leftarrow \text{Sigmoid}(\mathbf{W}_c \mathbf{h})$ \COMMENT{$[B, C]$}
\RETURN Multi-label predictions $\mathbf{y}$
\end{algorithmic}
\end{algorithm}

\begin{figure}[t]
\centering
\includegraphics[width=1.0\linewidth]{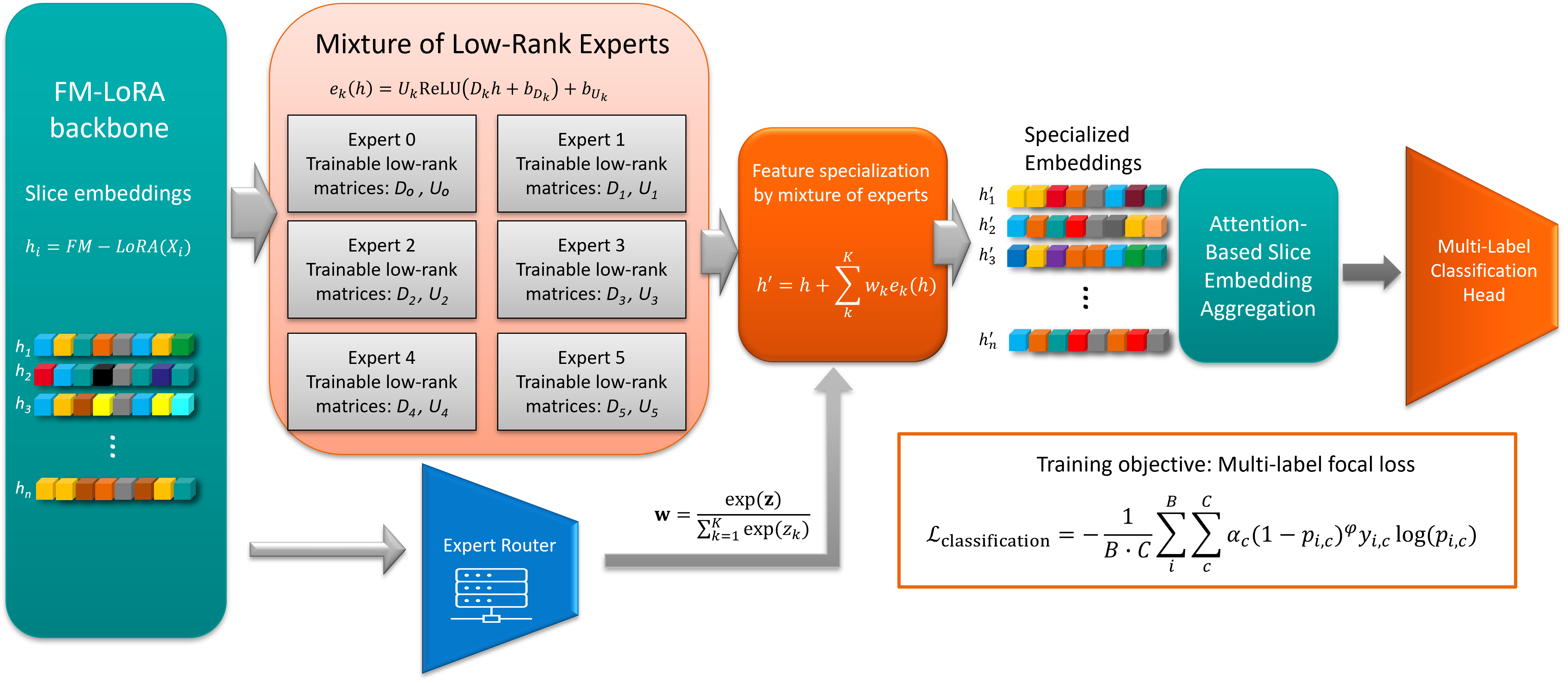}
\caption{\textbf{MoLRE: Parameter-efficient specialization for foundation models.} Illustrated for 2D models (e.g., DINOv3), MoLRE employs (1) multiple low-rank expert adapters for feature transformations, (2) an unsupervised soft router that learns to weight experts based on input features, and (3) attention-weighted pooling to aggregate slice-level features into volume-level representations. For 3D models (e.g., Pillar0-HeadCT), MoLRE is applied to spatially-pooled volumetric features, enabling conditional adaptation without explicit pathology supervision.}
\label{fig:method_diagram}
\end{figure}

For 3D models (e.g., Pillar0-HeadCT~\cite{agrawal2025pillar}) that natively process volumetric inputs, MoLRE is applied to spatially pooled volumetric features, with expert routing based on holistic volume-level representations rather than individual slices. The MoLRE formulation is identical across 2D and 3D models, differing only in the specialization level (slice-level for 2D, volume-level for 3D). Figure~\ref{fig:method_diagram} illustrates the 2D pipeline; for 3D models, MoLRE operates directly on the pooled volumetric embedding.

All models were trained end-to-end using a unified protocol. For 2D foundation models, we used multi-label focal loss~\cite{lin2017focalloss} with $\gamma=2.0$ and prevalence-based class weights $\alpha_t$, and applied parameter-efficient fine-tuning with LoRA (rank $r=8$, scaling $\alpha=16$) extended by MoLRE with $K=6$ low-rank experts (rank $r=8$) and router hidden dimension $d_h=256$. Optimization used AdamW~\cite{loshchilov2019decoupled} with learning rates of $10^{-3}$ for classification heads and $10^{-4}$ for LoRA/MoLRE parameters, alongside Repeat Factor Sampling~\cite{gupta2019lvis} to address rare findings. MoLRE routing was learned fully unsupervised via the task loss. For 3D foundation models, we followed the published training protocols. Models were trained for at least 20 epochs with early stopping after 5 epochs of no validation AUC improvement; results are reported from the best validation~checkpoint.

\section{Results and Discussion}

\subsection{Dataset and Preprocessing}
Anonymized non-contrast head CT (NCCT) volumes were retrospectively collected from nine centers across different geographical regions, with ethics approvals waiving informed consent. Scans were acquired on Siemens Healthineers, GE HealthCare, and Canon Medical Systems scanners. Exclusion criteria included age under 18 or missing axial reconstructions. In total, 72{,}756 NCCT studies met inclusion criteria, of which 65{,}542 were used for model development (58{,}251 for training and 7{,}291 for system optimization) and 7{,}214 were held out for independent evaluation. 

Leveraging recent advances in LLM-based medical content generation~\cite{bicknell2024chatgpt}, we automatically generated labels for 75 neurological findings from radiology reports using a private GPT-4-mini model deployed on a secure network. Finding categories were predefined by a neuroradiologist (Table~\ref{tab:pathology_categories}, Figure~\ref{fig:radar_plot}). To improve labeling quality, prompts were split into batches of 15 findings and restricted to the impression section to reduce false positives in follow-up studies. Label accuracy was validated on 90 cases annotated by a neuroradiologist, achieving an average of 0.64 false positives and 0.09 false negatives per case, with an overall finding-level accuracy of 0.986.

\begin{table}[t]
\centering
\caption{Pathology categories for the 75 head CT findings.}
\label{tab:pathology_categories}
\begin{tabular}{lcp{8cm}}
\toprule
\textbf{Category} & \textbf{Count} & \textbf{Example Findings} \\
\midrule
Hemorrhage & 12 & Basal ganglia hemorrhage, subdural hematoma, subarachnoid hemorrhage, epidural hematoma \\
Vascular & 8 & Acute infarct, chronic infarct, vessel occlusion, thrombosis \\
Trauma & 6 & Skull fractures, contusions, pneumocephalus, scalp hematoma \\
Mass/Lesion & 9 & Meningioma, cystic lesions, tumors, focal lesions \\
Structural & 11 & Hydrocephalus, herniation, midline shift, ventricular enlargement \\
Chronic/Other & 29 & Calcifications, atrophy, sinusitis, chronic changes \\
\bottomrule
\end{tabular}
\end{table}

Preprocessing involved automatic alignment of axial NCCT volumes to a standard reference frame, resampling to a 1-mm in-plane and 4-mm out-of-plane resolution, and normalization using HU windows: 0–80 HU, -20–180 HU, and -800–2000 HU, scaled to 0–1. Spatial augmentations include elastic deformation ($\alpha \in [0, 200]$, $\sigma \in [10, 13]$), small rotations ($\theta_x, \theta_y, \theta_z \in [-0.1, 0.1]$ radians), and anisotropic scaling (scale factor $\in [0.85, 1.15]$). Intensity augmentations consist of multiplicative brightness modulation ($m \in [0.99, 1.01]$) and additive Gaussian noise ($\mathcal{N}(0, \sigma^2)$ with $\sigma^2 \in [0, 0.03]$). Random axis-wise mirroring is applied with probability $p = 0.5$ for each axis independently.

\subsection{Multi-finding detection performance}

Table~\ref{tab:main_results_combined} summarizes multi-label CT finding detection performance across 75 pathology classes. MoLRE consistently improves all compatible foundation models with less than 0.5\% additional parameters; DeepCNTD~\cite{yoo2025non} is excluded due to the absence of attention pooling. Absolute AUC gains range from $+0.2\%$ to $+4.6\%$, with MedGemma+MoLRE achieving the highest performance at 0.917 AUC. Improvements are larger for smaller or less specialized models (e.g., DINOv3-Base, $+4.6\%$) than for larger-capacity architectures (e.g., DINOv3-Large, $+0.3\%$), indicating that mixture-of-experts routing is most beneficial when base model capacity is limited.

\begin{table}[t]
\centering
\caption{\textbf{Multi-label head CT detection performance and model complexity.} Mean AUC $\pm$ std across 75 neurological findings and parameter counts for each foundation model. All 2D models use LoRA fine-tuning with attention-based slice aggregation. Pillar0-HeadCT is fully fine-tuned. DeepCNTD is excluded from MoLRE evaluation due to the absence of attention pooling.}
\label{tab:main_results_combined}
\small
\setlength{\tabcolsep}{4pt}
\begin{tabular}{@{}llcccccc@{}}
\toprule
Model & Domain & Total & Base AUC & +MoLRE & Trainable & MoLRE \\
 &  & Params &  & AUC & Params & Params \\
\midrule
\multicolumn{7}{l}{\textit{General 2D Vision Models}} \\
DINOv3-Base  & Natural & 86.6M  & 0.856$\pm$0.083 & \textbf{0.902}$\pm$0.054 & 0.59M & 0.28M \\
DINOv3-Large & Natural & 305.2M & 0.910$\pm$0.052 & \textbf{0.913}$\pm$0.051 & 1.18M & 0.37M \\
\midrule
\multicolumn{7}{l}{\textit{Medical 2D Vision-Language Models}} \\
MedImageInsight & Medical & 364.3M & 0.863$\pm$0.080 & \textbf{0.876}$\pm$0.081 & 1.42M & 0.37M \\
MedGemma        & Medical & 431.1M & 0.874$\pm$0.084 & \textbf{0.917}$\pm$0.049 & 1.68M & 0.41M \\
\midrule
\multicolumn{7}{l}{\textit{Domain-Specific 3D Models}} \\
Pillar0-HeadCT & Head CT & 113.9M & 0.891$\pm$0.067 & \textbf{0.893}$\pm$0.062 & 113.9M & 0.41M \\
DeepCNTD       & Head CT & 6.9M   & 0.832$\pm$0.090 & -- & 1.08M & -- \\
\bottomrule
\end{tabular}
\end{table}

MedImageInsight is trained with a discriminative objective that explicitly separates medical concepts in embedding space by maximizing agreement between paired image–text samples~\cite{radford2021learning} and minimizing similarity across mismatched pairs. In contrast, MedGemma’s vision encoder is optimized within a generative framework to produce features that support language model outputs for tasks such as visual question answering and report generation. Although MedGemma processes image–text pairs during pre-training, its optimization target emphasizes generative likelihood rather than discriminative embedding structure, yielding features that are semantically rich and text-aligned but less optimally organized for pure visual classification. The substantial gains achieved by MoLRE indicate that explicit specialization can recover task-specific discrimination that is underutilized in generatively optimized features. Notably, MedGemma attains strong base performance, suggesting that generative pre-training provides broad semantic understanding, but this advantage is more fully realized only after conditional routing is applied.

MoLRE yields smaller gains for the 3D volumetric model Pillar0-HeadCT than for 2D slice-based foundation models, indicating that conditional expert routing depends on feature heterogeneity at the point of specialization. In 2D architectures, slice embeddings encode spatially localized anatomical content with distinct pathology distributions, enabling effective slice-level expert specialization. In contrast, Pillar0-HeadCT applies progressive spatial pooling that collapses this heterogeneity into a single volume-level representation before MoLRE is applied, removing the spatial structure that facilitates pathology-specific routing. This does not diminish the value of 3D models, as Pillar0-HeadCT’s strong base performance (0.891 AUC) reflects the benefits of 3D spatial inductive biases, but rather highlights a trade-off: 3D models capture global volumetric context, while 2D models preserve the spatial granularity required for location-specific conditional specialization.

\subsection{Stratified Detection and Per-Finding Performance}
\begin{figure}[t]
\centering
\includegraphics[width=1.0\linewidth]{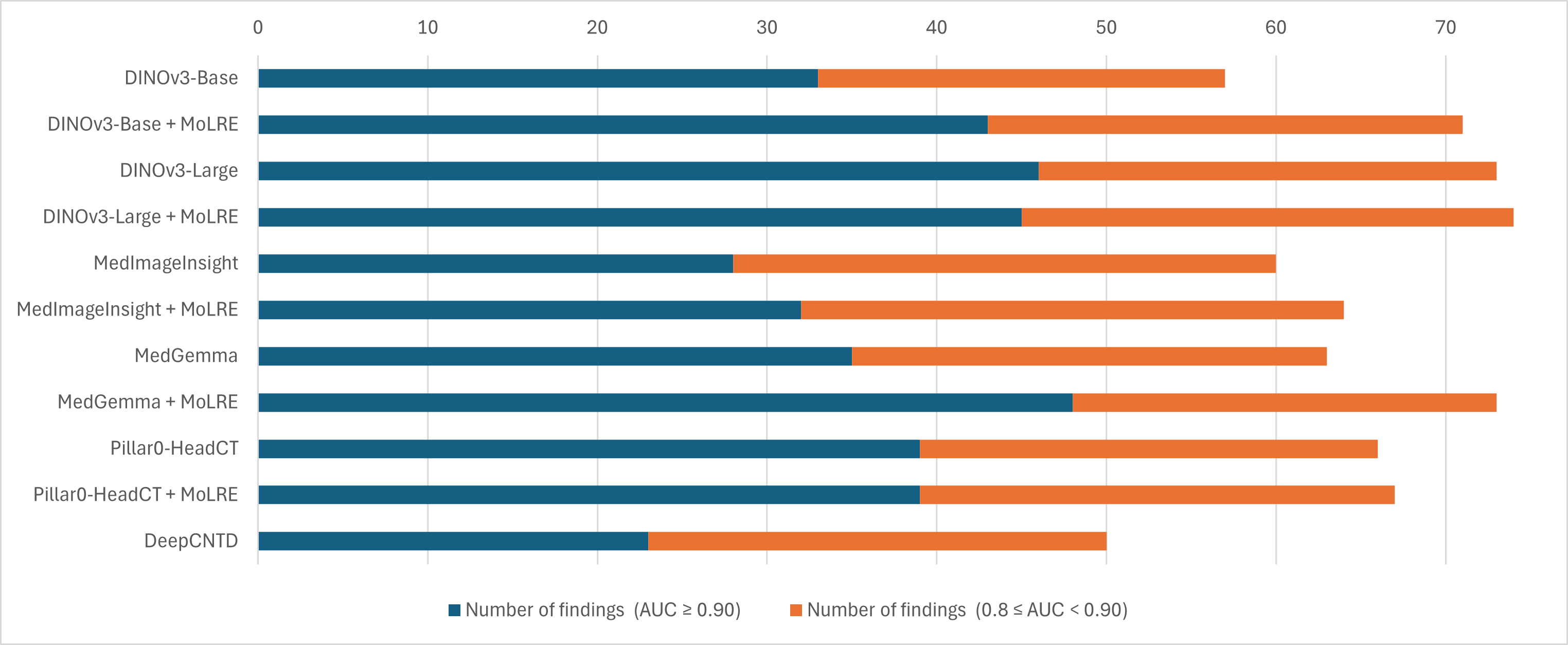}
\caption{\textbf{Stratified multi-finding detection performance.} Numbers of neurological findings achieving high-confidence performance (AUC $\geq 0.90$) and moderate performance ($0.8 \leq \text{AUC} < 0.9$) for each foundation model, with and without MoLRE.}
\label{fig:stratified_perf}
\end{figure}

Figure~\ref{fig:stratified_perf} and the accompanying stratified analysis reveal that MoLRE primarily improves performance by shifting a larger number of findings into the high-confidence regime (AUC $\geq 0.90$), rather than uniformly increasing mid-range performance. For DINOv3-Base, MoLRE increases the number of high-performing findings from 33 to 43, indicating that conditional expert routing substantially sharpens detection for many findings that were previously near—but below—the high-AUC threshold. A similar pattern is observed for MedGemma, where the number of findings with AUC $\geq 0.90$ increases markedly from 35 to 48, accompanied by a reduction in the $0.8 \leq \text{AUC} < 0.9$ range, suggesting that MoLRE effectively promotes moderately strong detectors into a highly reliable regime. In contrast, larger or more specialized models exhibit diminishing returns: DINOv3-Large shows a redistribution within the high- and mid-AUC ranges rather than a net gain, and Pillar0-HeadCT demonstrates minimal stratified change, consistent with its limited overall MoLRE benefit. MedImageInsight shows modest gains concentrated at the high-AUC threshold without altering the mid-range distribution.

\begin{figure}[t]
\centering
\includegraphics[width=1.0\linewidth]{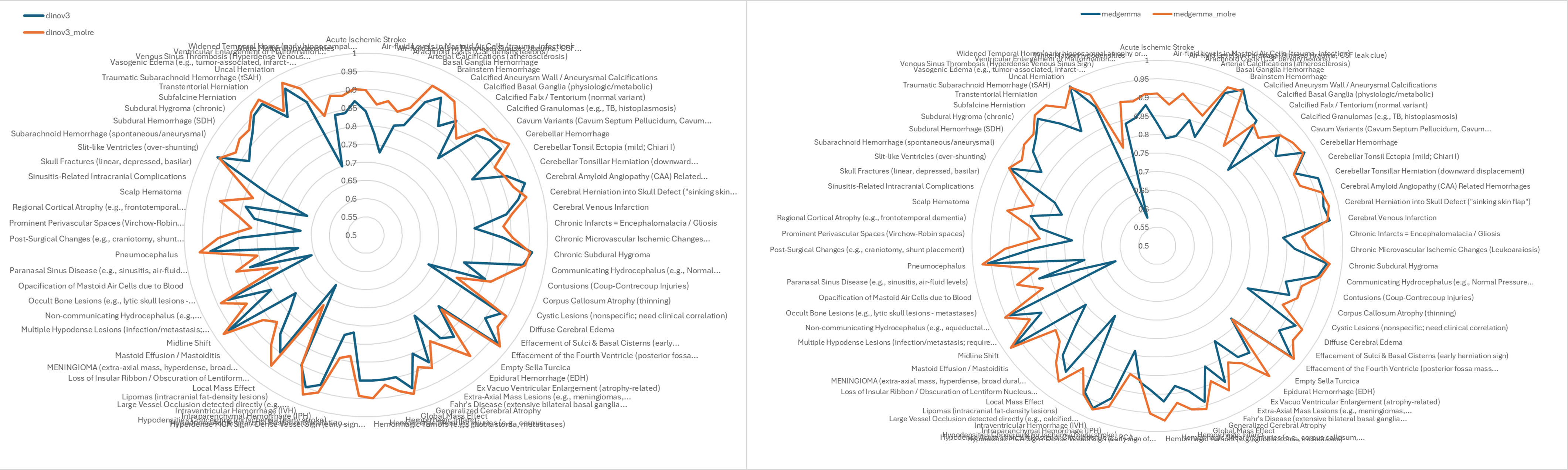}
\caption{\textbf{Per-finding detection performance with and without MoLRE.} Radar plots comparing baseline models and their MoLRE-enhanced counterparts for \textbf{left:} DINOv3-Base vs.\ DINOv3-Base+MoLRE and \textbf{right:} MedGemma vs.\ MedGemma+MoLRE across 75 neurological findings.}
\label{fig:radar_plot}
\end{figure}

Figure~\ref{fig:radar_plot} presents per-finding AUC comparisons between baseline models and their MoLRE-enhanced counterparts, revealing that MoLRE delivers the largest gains for findings that are visually subtle, heterogeneous, or underrepresented. For DINOv3-Base, MoLRE consistently improves detection across a wide range of ischemic, traumatic, and mass-related findings, with particularly pronounced gains for early ischemic signs (e.g., loss of insular ribbon, hypodense acute infarcts), extra-axial and cystic lesions, skull and mastoid-related trauma, and vascular abnormalities such as venous sinus thrombosis. These findings typically exhibit lower baseline AUCs and benefit from conditional specialization that sharpens sensitivity to localized or low-contrast imaging cues. MedGemma exhibits a similar but more selective pattern: while baseline performance is already strong for many hemorrhagic and mass-effect findings, MoLRE substantially boosts performance for rare or complex entities such as lipomas, occult bone lesions, venous sinus thrombosis, and sinus-related intracranial complications, often yielding large absolute AUC improvements. In contrast, findings with very high baseline performance—such as major hemorrhage types, ventricular enlargement, and herniation syndromes—show minimal change or mild saturation effects, indicating limited headroom for improvement. Overall, the radar plots demonstrate that MoLRE primarily enhances performance where discriminative features are weak or fragmented in the base representation, reinforcing its role as a targeted specialization mechanism rather than a uniform performance booster.

\section{Conclusion}
In this work, we presented a comprehensive benchmark of Mixture of Low-Rank Experts across diverse medical imaging foundation models on large-scale, multi-label head CT analysis with 75 neurological findings. Our results demonstrate that MoLRE provides consistent benefits across models, with absolute AUC gains ranging from 0.2\% to 4.6\%, revealing that adaptation benefit depends on a complex interaction between pre-training domain and strategy, model architecture, and model size, rather than on model scale alone. By enabling conditional, parameter-efficient specialization, MoLRE allows general-purpose and medical-general foundation models to substantially improve performance, while task-specific or volumetric architectures exhibit diminishing but non-zero gains. Notably, MedGemma equipped with MoLRE achieves state-of-the-art performance with an average AUC of 0.917, showing that lightweight expert routing can match or exceed the effectiveness of full fine-tuning. Beyond absolute performance, our stratified and per-finding analyses show that MoLRE primarily consolidates performance among challenging or borderline findings, elevating many into a high-confidence detection regime. Together, these findings provide new empirical insight into when and why conditional adaptation is effective, highlighting MoLRE as a practical and scalable strategy for specializing foundation models in complex clinical imaging tasks, particularly in resource-constrained deployment settings.

\subsubsection{Disclaimer}
The concepts and information presented in this paper are based on research results that are not commercially available. Future availability cannot be guaranteed.

\bibliographystyle{splncs04}
\bibliography{ref}

\end{document}